\documentclass[journal,twoside,web]{ieeecolor}

\usepackage{tmi}
\usepackage{amsmath,amssymb,amsfonts}
\usepackage{graphicx}
\usepackage{float}

\makeatletter
\let\orig@IEEEmakecaption\@makecaption
\long\def\@makecaption#1#2{%
  \ifx\@captype\@IEEEtablestring
    \orig@IEEEmakecaption{#1}{#2}%
  \else
    \@IEEEfigurecaptionsepspace
    \setbox\@tempboxa\hbox{\footnotesize Fig.~\thefigure.~~\sf #2}%
    \ifdim \wd\@tempboxa >\hsize
      {\footnotesize\noindent{#1.}~~{\sf #2}\par}%
    \else
      \ifcenterfigcaptions
        \hbox to\hsize{\footnotesize\hfil{#1.}~~{\sf #2}\hfil}%
      \else
        \hbox to\hsize{\footnotesize{#1.}~~{\sf #2}\hfil}%
      \fi
    \fi
  \fi}
\makeatother
\usepackage{textcomp}
\usepackage{array}
\usepackage{multirow}
\usepackage{booktabs}
\makeatletter\let\NAT@parse\undefined\makeatother
\usepackage{hyperref}
\graphicspath{{./figures/}}
\def\BibTeX{{\rm B\kern-.05em{\sc i\kern-.025em b}\kern-.08em
    T\kern-.1667em\lower.7ex\hbox{E}\kern-.125emX}}

\makeatletter
\def\ps@headings{%
  \def\@oddhead{}\def\@evenhead{}%
  \def\@oddfoot{}\def\@evenfoot{}%
}
\def\ps@titlepagestyle{%
  \def\@oddhead{}\def\@evenhead{}%
  \def\@oddfoot{}\def\@evenfoot{}%
}
\makeatother
\begin{document}

\title{Task-Based CT Protocol Optimization Using Reinforcement Learning and Virtual Imaging Trials}

\author{
Jiaqi~Zou$^*$, \and
David~Fenwick$^*$, \and
Vahid~Tarokh, \and
Nicholas~Felice, \and
Jayasai~Rajagopal, \and
Anuj~Kapadia, \and
Ehsan~Samei, \and
Navid~NaderiAlizadeh$^\dagger$, \and
Ehsan~Abadi$^\dagger$
\thanks{$^*$Equal Contribution.}
\thanks{$^\dagger$Equal supervision.}
\thanks{Manuscript received August XX, 2026. This work was supported in part by the National Institutes of Health (R01HL155293 and P41EB028744). This manuscript has been authored in part by UT-Battelle, LLC, under contract DE-AC05-00OR22725 with the US Department of Energy (DOE). The US government retains and the publisher, by accepting the article for publication, acknowledges that the US government retains a nonexclusive, paid-up, irrevocable, worldwide license to publish or reproduce the published form of this manuscript, or allow others to do so, for US government purposes. DOE will provide public access to these results of federally sponsored research in accordance with the DOE Public Access Plan (\href{http://energy.gov/downloads/doe-public-access-plan}{http://energy.gov/downloads/doe-public-access-plan}). (\textit{Corresponding author: Jiaqi Zou}.)}
\thanks{J. Zou is with the Department of Electrical and Computer Engineering, Duke University, Durham, NC 27705 USA (e-mail: \href{mailto:jiaqi.zou@duke.edu}{jiaqi.zou@duke.edu}).}
\thanks{D. Fenwick and N. Felice are with the Department of Radiology and Medical Physics Graduate Program, Duke University, Durham, NC 27705 USA (e-mail: \href{mailto:david.fenwick@duke.edu}{david.fenwick@duke.edu}). }
\thanks{V. Tarokh is with the Department of Electrical and Computer Engineering, Duke University, Durham, NC 27705 USA.}
\thanks{J. Rajagopal and A. Kapadia are with Oak Ridge National Laboratory, Oak Ridge, TN 37830 USA.}
\thanks{E. Samei is with the Departments of Electrical and Computer Engineering, Radiology, Biomedical Engineering, Physics, and Medical Physics Graduate Program, Duke University, Durham, NC 27705 USA.}
\thanks{N. NaderiAlizadeh is with the Department of Biostatistics and Bioinformatics, Duke University, Durham, NC 27705 USA (e-mail: \href{mailto:navid.naderi@duke.edu}{navid.naderi@duke.edu}).}
\thanks{E. Abadi is with the Departments of Electrical and Computer Engineering, Radiology, and Medical Physics Graduate Program, Duke University, Durham, NC 27705 USA (e-mail: \href{mailto:ehsan.abadi@duke.edu}{ehsan.abadi@duke.edu}).}}
\maketitle

\begin{abstract}
Protocol optimization in computed tomography (CT) seeks to achieve high diagnostic image quality while reducing radiation dose. However, the interdependence of acquisition and reconstruction parameters makes exhaustive protocol testing impractical. To address this challenge, we propose a virtual imaging trial (VIT) framework with reinforcement learning to optimize CT protocols more accurately and efficiently. Sixty-three computational human models with liver lesions were imaged with a validated CT simulator across 468 combinations of acquisition and reconstruction parameters, including tube voltage, tube current, reconstruction kernel, slice thickness, and pixel size. The objective of the optimization task was to maximize liver lesion detection, quantified by detectability index ({\boldmath$d'$}), while reducing the radiation dose (mAs). With this objective, a Proximal Policy Optimization (PPO) agent was trained. To inform the PPO agent with patient attributes, their CT localizers were represented via compact embeddings derived from a pre-trained vision transformer. On held-out patients, the framework approached the exhaustive-search objective at a small fraction of its cost: simulating only 8 protocols per patient---about 2\% of the 468 used for exhaustive testing while recovering 98.2\% of the oracle (seed SD 1.5; patient-level 95\% CI 96.6--99.2); with no patient-specific simulation, the framework had a recovery of 89.7\% (seed SD 0.5; patient-level 95\% CI 80.9--95.7) from surrogate scoring alone. Conditioning on the localizer significantly improved recovery: with no patient-specific simulation, the localizer-conditioned policy exceeded its localizer-blind counterpart by 10.7 percentage points (paired 95\% CI 2.9--19.5; \boldmath$p=0.02$). Our results demonstrate that the proposed framework can eliminate the need for exhaustive protocol testing at deployment, providing a practical route to task-based, dose-aware protocol selection prior to the diagnostic scan.
\end{abstract}

\begin{IEEEkeywords}
Computed tomography, optimization, reinforcement learning, vision transformers, virtual imaging trials.
\end{IEEEkeywords}

\section{Introduction}
\label{sec:introduction}
\IEEEPARstart{C}{omputed} tomography (CT) protocols specify the acquisition and reconstruction parameters that collectively define how CT images are acquired and processed \cite{RN1}. These protocols must be optimized to achieve high diagnostic image quality while keeping radiation dose as low as possible. A conventional approach to CT optimization involves acquiring images across a wide range of parameter settings and evaluating their impact on diagnostic image quality. However, such repeated scans are not feasible in human subjects due to added radiation exposure. Alternatively, these repeated scans can be acquired and analyzed using physical phantoms \cite{RN2,RN3}. While this approach is safer and more practical, physical phantoms lack the anatomical and pathological variability of real patients, which limits their clinical applicability. Furthermore, the number of possible parameter combinations often makes it infeasible to test all configurations exhaustively. This leads to limited sampling of the parameter space and increases the risk of converging to suboptimal protocols.

These limitations can be addressed using realistic virtual imaging trials (VITs) \cite{RN14}, in which a conventional imaging trial is simulated using validated human and scanner models. VITs enable controlled imaging experiments across a wide range of imaging conditions without exposing patients to radiation, while also providing anatomical and physiological ground-truth data. Accordingly, VITs have been widely used for image quality assessment and optimization \cite{RN11, RN12, RN13}.

For instance, Barufaldi \textit{et al.} used VITs to evaluate and optimize a prototype for simultaneous Digital Breast Tomosynthesis (DBT) and Mechanical Imaging\cite{RN11}. Similarly, Vancoillie \textit{et al.} used VITs to study microcalcification detectability in DBT and synthetic 2D mammography across different acquisition parameters\cite{RN12}. In another study, Abadi \textit{et al.} employed a VIT approach to test various beam collimation and pitch settings to determine their impact on image quality attributes for CT images with respiratory and cardiac motion\cite{RN13}. 

\begin{figure}[t]
\centering
\includegraphics[width=\linewidth]{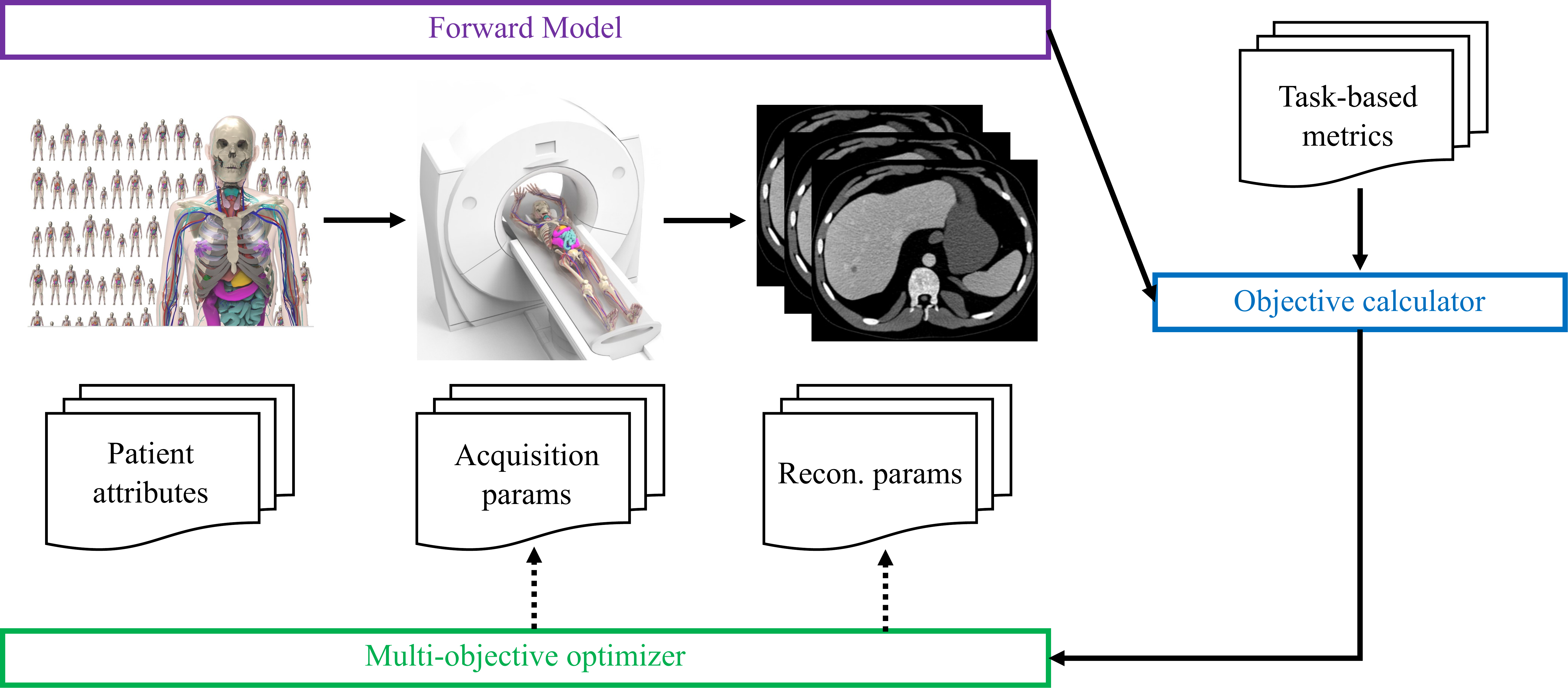}
\caption{CT protocol optimization framework using virtual imaging trials (VITs) as the forward model.}
\label{fig:vit-concept}
\end{figure}

While these efforts have provided valuable insights, they have been limited by the need to study a reduced set of imaging parameters rather than conducting an exhaustive search, primarily due to the high computational demands associated with acquiring and reconstructing images across a larger parameter space. The purpose of this study is to address this limitation by developing an optimization framework that combines VITs with reinforcement learning (Fig.~\ref{fig:vit-concept}), enabling efficient and accurate CT protocol optimization beyond limited parameter sampling. The proposed framework is benchmarked for its effectiveness in optimizing CT protocols for liver lesion detection. 

\section{Methods}
\label{sec:methods}

\subsection{Study Framework}
\begin{figure}[!t]
\centering
\includegraphics[width=\linewidth]{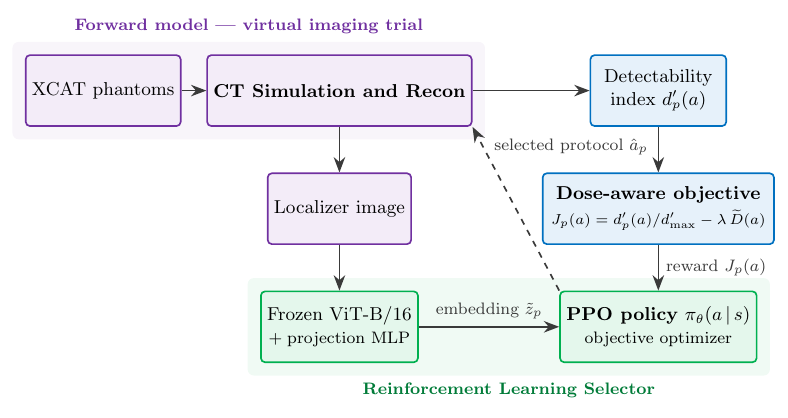}
\caption{Overview of the proposed framework. The forward model generates CT images of human models for any protocol acquisition-reconstruction setting, denoted by $a$. It also generates the patient's CT localizer image. The objective ($J_p(a)$) balances image quality (detectability index $d'_p(a)$) against radiation dose (tube current $D(a)$), with a fixed weight $\lambda$ controlling the trade-off. The subscript $p$ denotes the patient. ViT is vision transformer and MLP is multi-layer perceptron. The PPO policy, conditioned on the localizer embedding $\tilde z_p$ and rewarded by $J_p(a)$, returns its selected protocol $\hat{a}_p$ to the simulator, closing the optimization loop.}
\label{fig:framework}
\end{figure}

This study developed a framework that integrates VITs with reinforcement learning to optimize CT imaging protocols. An overview of the framework is shown in Fig.~\ref{fig:framework}. In summary, computational human models with liver lesions were imaged using a validated CT scanner model, and the resulting sinograms were reconstructed using an open-source reconstruction toolkit. The reconstructed images were analyzed to estimate spatial resolution, noise magnitude, and noise texture, which were used to compute the lesion detectability index ($d'$). A dose-aware objective was set to balance $d'$ maximization against a radiation dose penalty. A reinforcement-learning agent served as the protocol selector, using the patient's localizer image to select a protocol for each patient. The performance was evaluated on held-out patients against an exhaustive-search oracle.

\subsection{Computational Phantoms with Liver Lesions}
Extended cardiac-torso (XCAT) phantoms represented a cohort of 63 patients with liver lesions (35 male, 28 female; mean body mass index (BMI) of 26.8\,$\pm$\,5.5~kg/m$^2$, and BMI range 15.5--38.8~kg/m$^2$), spanning diverse body habitus and lesion characteristics \cite{RN15}. Each model included detailed liver anatomy and hepatic vasculature generated using a physics-based vessel growth algorithm \cite{RN16}. One to six liver lesions of varying sizes were placed randomly within each liver. Details of these models are provided in \cite{RN15}.

\subsection{Image Acquisition}
The human models were imaged using a validated CT simulator (DukeSim) to generate projection images \cite{RN4, RN19, RN20, RN21}. DukeSim was configured as a vendor-generic scanner (Duke-1) whose geometry and physics represent those of modern CT systems \cite{RN4}. The acquisition parameter space included tube voltages (kV) and tube currents (mAs), which are commonly used in clinical practice, as summarized in Table \ref{tab:parameters}.

Due to the large number of simulations required for this study, the acquisitions were performed at Oak Ridge National Laboratory using the Summit supercomputer, which had 4600 compute nodes, each with six NVIDIA Tesla V100 GPUs, two IBM POWER9 CPUs, and 1~TB of coherent memory \cite{RN27}.

\subsection{Image Reconstruction}
The acquired CT sinograms were reconstructed using the open-source Multi-Channel Reconstruction Toolkit (MCR Toolkit 2) \cite{RN5}. The reconstructions used a filtered back-projection technique in which the filter shape is defined by the filter kernel and the filter $f_{50}$, setting image sharpness. The reconstruction parameters studied are summarized in Table \ref{tab:parameters}. Reconstructed pixel sizes of 0.5 and 1.0~mm correspond to 1024 and 512 reconstruction matrices, respectively, at a fixed reconstruction field of view. 

\begin{table}[ht]
\caption{Acquisition and reconstruction parameters}
\label{tab:parameters}
\centering

\renewcommand{\arraystretch}{1.3}
\setlength{\tabcolsep}{6pt}

\begin{tabular}{@{}>{\raggedright\arraybackslash}p{58pt}
                >{\raggedright\arraybackslash}p{68pt}
                >{\raggedright\arraybackslash}p{90pt}@{}}
\toprule
\textbf{Parameter Type} &
\textbf{Parameter Name} &
\textbf{Parameter Space} \\
\midrule
\multirow[t]{2}{58pt}{Acquisition} &
Tube Voltage &
100, 120, 140 kV \\
& Tube Current &
25, 80, 150 mAs \\
\midrule
\multirow[t]{4}{58pt}{Reconstruction} &
Filter Kernel &
Ram-lak, Cosine, Smooth, Sharp, Enhancing \\
& Filter $f_{50}$ &
0.4, 0.6, 0.8 mm$^{-1}$ \\
& Slice Thickness &
0.5, 1.0 mm \\
& Pixel Size &
0.5, 1.0 mm \\
\bottomrule
\end{tabular}

\vspace{2pt}
{\footnotesize\begin{minipage}{\columnwidth}\raggedright
\textit{Notes.} Not all kernel-$f_{50}$ combinations are defined; $13$ of $15$ kernel-$f_{50}$ pairs were used (the remaining two pairs are not defined by the reconstruction toolkit), giving $3 \times 3 \times 13 \times 2 \times 2 = 468$ protocols. Kernel and $f_{50}$ are therefore treated as a single $13$-level axis. The same 468-protocol space was used for the surrogate, all learned selectors, and all baselines.
\end{minipage}}
\end{table}

\subsection{Task-Based Image Quality}
The task-based image quality objective was defined as lesion detection and quantified using detectability index \(d'\), a metric that reflects the likelihood of a lesion detected by a human reader. \(d'\) was calculated in accordance with Smith \textit{et al.} \cite{RN26}:
\begin{align}
    d'^2=\frac{[\iint|W(u,v)|^2\cdot MTF^2(u,v)du\,dv]^2}{\iint|W(u,v)|^2\cdot MTF^2(u,v)\cdot NPS(u,v)du\,dv},\end{align}
where \(W(u,v)\) is the Fourier transform of the task function defined by the lesion size and image contrast, \(MTF(u,v)\) is the modulation transfer function, and \(NPS(u,v)\) is the noise power spectrum. Since the actual lesion size and image contrast are not available prior to the scan, we defined the task function as a hypothetical spherical lesion with a fixed diameter of 4.2 mm and image contrast of 43.8 HU, reflecting the average lesion size and image contrast in our library.

MTF was estimated directly from the reconstructed images \cite{RN22}: the skin-to-air interface was segmented to estimate the edge-spread function (ESF), the ESF was differentiated to obtain the line-spread function (LSF) and fit to the model of \cite{RN23}, and the Fourier transform of the fitted LSF yielded MTF.

NPS was estimated by selecting 100 random soft-tissue ROIs with mean CT numbers between \(-300\) and \(300\) Hounsfield units (HU), excluding air, bone, skin, and lesion voxels. Within each ROI, a second-order polynomial was fitted and subtracted to remove low-frequency background non-uniformity, leaving the noise-only residual. The two-dimensional NPS was then computed as the ensemble average, over all ROIs, of the squared magnitude of the discrete Fourier transform of the residual, scaled by the pixel area and ROI dimensions, and was radially averaged to yield the one-dimensional NPS as a function of spatial frequency \cite{RN24}.

\subsection{Dose-Aware Objective}
For each patient $p$, the performance of an imaging protocol $a$ was evaluated using a dose-aware objective:
\begin{align}
J_p(a)
=
\frac{d'_{p}(a)}{d'_{\mathrm{max}}}
-
\lambda\,
\frac{D(a)-D_{\min}}{D_{\max}-D_{\min}},
\label{eq:dose_aware_objective}
\end{align}
where $d'_{p}(a)$ is the detectability index for patient $p$ at protocol $a$, normalized by $d'_{\mathrm{max}}$, which is the maximum lesion detectability over the training dataset used in this study. $D(a)$ is a tube-current (mAs) value that was used as a dose surrogate, normalized over the range of mAs values used in this study ($D_{\min}=25$ and $D_{\max}=150$~mAs). The weight $\lambda$ controls the detectability--dose trade-off, with larger values favoring lower dose and smaller values favoring higher detectability. 

While we report results on the full search space we considered for this parameter, i.e., $\lambda\in\{0.14,0.16,0.18,0.20\}$ (see Fig.~\ref{fig:lambda_sensitivity}, computed on the training patients only), we use $\lambda=0.18$ as a balanced operating point for the main comparisons.

We formulate the protocol optimization problem as follows, with the solution representing the oracle protocol: 
\begin{align}
a_p^\star
=
\arg\max_{a \in \mathcal{A}} J_p(a),
\label{eq:oracle}
\end{align}
where $\mathcal{A}$ denotes the full protocol action space. The oracle protocol was computed via exhaustive search and used as an evaluation reference.

\begin{figure}[!t]
\centering
\includegraphics[width=\linewidth]{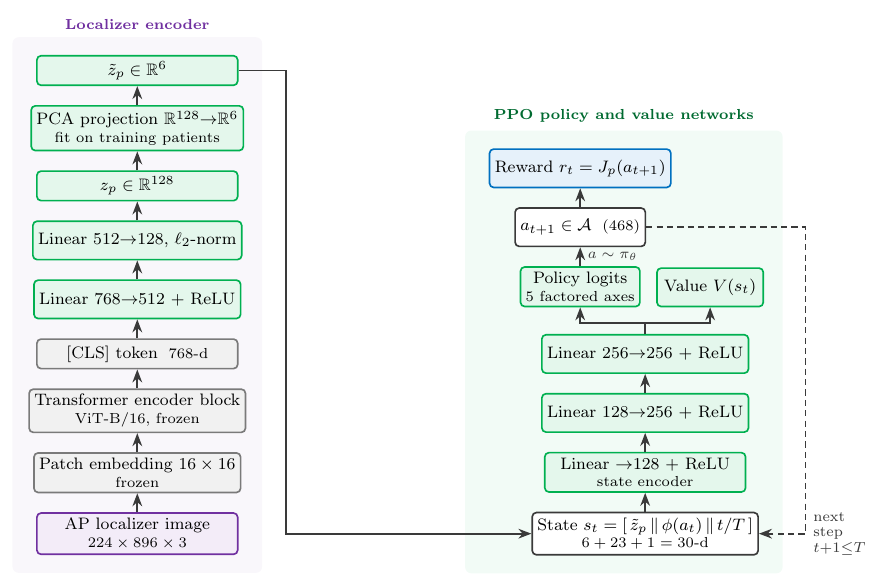}
\caption{Architectures and training of the localizer-conditioned selector (colors as in Fig.~\ref{fig:vit-concept}; gray: frozen; green: trainable). The ViT-B/16 backbone is frozen; only the projection MLP is trained, with a within-patient protocol-ranking loss on the training patients so that the embedding orders protocols by $d'$ within each patient, and the PCA projection is fit on the training patients only. The policy and value networks are trained with the clipped PPO objective ($\gamma=0.99$) under the eight-step episode budget, with the dose-aware objective $J_p$ as the per-step reward. At deployment, each visited protocol is scored by the random-forest surrogate and the best-seen protocol $\hat{a}_p$ is committed; the off-line mode does not have access to held-out ground truth.}

\label{fig:rl_architecture}
\end{figure}

\subsection{Localizer Encoding via Vision Transformers}
To personalize protocol selection before the diagnostic scan, the policy must be conditioned on information available about the individual patient at that time. The anteroposterior (AP) localizer image is a patient image acquired before the diagnostic scan in the routine CT workflow, and it captures the body size, shape, and anatomy. We leveraged AP localizer images to extract compact, patient-specific representations for the developed RL agent (Fig.~\ref{fig:rl_architecture}).

Localizer images were generated for each human model with the same validated CT simulator (DukeSim) used for the tomographic acquisitions, yielding an attenuation line-integral radiograph with a native size of $224\times 900$ pixels. Each localizer was normalized by a robust (1\textsuperscript{st}--99\textsuperscript{th} percentile) intensity stretch, resized with preserved aspect ratio and letterbox padding to $224\times896$ pixels.

Because our cohort was too small to train an image encoder from scratch without overfitting, we used a Vision Transformer \cite{dosovitskiy2021vit} pretrained on natural images as a fixed feature extractor. The Vision Transformer operated on the grayscale localizer images, replicated to the three input channels expected by the model. The localizer's short dimension, $224$, matches the native input size of the Vision Transformer so that its $16\times16$ patch embedding operates at the scale it was pretrained on, and the long dimension, $896=4\times224$, preserves the localizer's approximately $1{:}4$ aspect ratio without distortion. This input produces a $14\times56$ grid of patch tokens rather than the pretrained $14\times14$ grid, so the encoder's positional embeddings were interpolated from $14\times14$ to $14\times56$.

For a given patient's localizer image, the Vision Transformer generated a $768$-d CLS-token embedding at its output. We then trained a small projection multi-layer perceptron (MLP) on top of this embedding. More specifically, the CLS-token embedding was passed through a two-layer perceptron ($768\!\rightarrow\!512\!\rightarrow\!128$ with rectified linear unit (ReLU) activation) and $\ell_2$-normalized to yield the final embedding for each patient $p$, denoted by $z_p$.

The projection MLP was optimized on the training patients only, using an auxiliary predictor that ranks, for each patient, the protocols with an available simulation by their $d'$; the loss combined a Huber term on the normalized detectability, a within-patient pairwise-ranking term, and an augmentation-consistency term. As a result, $z_p$ captures the patient-specific information relevant to protocol ranking. Optimization used Adam (learning rate $3\times10^{-4}$, weight decay $10^{-5}$, batch size $64$) with early stopping on a validation split of the training patients.

We finally project $z_p$ onto its first six principal components (PCA-6) and denote the resulting low-dimensional embedding by $\tilde{z}_p$. The PCA projection was fit only to the training patients; these six components capture $93\%$ of the training-set embedding variance while discarding patient-level variation unrelated to protocol ranking. We use these six components for all downstream selectors; the comparison against the full $128$-dimensional embedding is reported in Section~\ref{sec:results}. The embedding dimension was selected based on the training patients only: five-fold cross-validation of the surrogate selector across candidate dimensions ($2$--$10$) showed recovery rising sharply through about four components and then plateauing, and we retained six components within this plateau.

The patient cohort was split into 51 training patients and 12 held-out test patients. The held-out test patients were strictly excluded from the localizer-embedding training, surrogate fitting, policy optimization, hyperparameter selection, and baseline training. The test set consisted only of patients with complete 468-protocol objective tables, enabling a fully simulated exhaustive-search oracle for each held-out patient. The normalization constant $d'_{\mathrm{max}}$ and the operating point $\lambda$ were likewise fixed using the training patients only.

\subsection{Data Imputation with a Random-Forest Surrogate}
\label{sec:rf}
It is computationally expensive to generate CT images for training patient across all protocol combinations. As such, some combinations were not simulated. Of the $51\times468=23{,}868$ training combinations, $85.9\%$ ($20{,}505$) were completed and the remaining conditions ($14.1\%$; $3{,}363$) were missing. These missing conditions were completed by training a random-forest surrogate that predicts $d'$ for any patient protocol pair. The same trained surrogate was reused later by the protocol selectors (see Sections~\ref{sec:ppo} and~\ref{sec:baselines}).

The surrogate takes two inputs: the scan protocol and the patient's six-dimensional localizer embedding $\tilde z_p$. The protocol is encoded axis by axis (Table~\ref{tab:parameters}). Tube potential, tube current, slice thickness, and $f_{50}$ are ordinal, so they enter as numeric values; this lets the trees split on thresholds and interpolate to dose and sharpness levels that were never simulated, which is what imputation requires. The reconstruction kernel has no natural order and the pixel size has only two levels, so both enter as one-hot indicators. The policy state of Section~\ref{sec:ppo} instead encodes every axis as one-hot, because the policy chooses among discrete levels and never needs to interpolate; the two encodings differ because the two tasks differ. Each forest used $400$ trees (maximum depth $20$, minimum leaf size $2$) and was fit on the simulated cells of the training patients only. All twelve held-out test patients retained the complete $468$-protocol simulation, used only for oracle evaluation and never for surrogate fitting.

The surrogate is used only to \emph{select} protocols; every reported recovery is computed from the fully simulated exhaustive-search oracle and not from surrogate predictions. We report its predictive fidelity in Section~\ref{sec:results}. Further, as an ablation study, a localizer-blind variant was also retained by dropping the embedding and using the protocol features alone.

\subsection{Protocol Selectors}
\label{sec:selectors}

\subsubsection{Localizer-Conditioned PPO Selector}
\label{sec:ppo}
We cast protocol selection as a budgeted search solved by reinforcement learning (RL) over $T$ steps. At step $t\in\{0,\dots,T-1\}$, the input to the PPO agent, i.e., the environment state in RL terminology, consists of a concatenation of the PCA-reduced localizer embedding, a factored (per-axis) one-hot encoding of the currently selected protocol, and the normalized step index, i.e., $s_t=[\,\tilde z_p \,\Vert\, \phi(a_t)\,\Vert\, t/T\,] \in \mathbb{R}^{30}$, where $\tilde z_p\in\mathbb{R}^{6}$ is the PCA-reduced localizer embedding and $\phi(a_t)\in\{0,1\}^{23}$ stacks the per-axis one-hot codes of the five protocol axes ($3+3+13+2+2 = 23$ levels). The state thus encodes the patient, the currently selected protocol, and the remaining budget. The action represents a complete protocol choice, i.e., a factored discrete choice over the five protocol axes (tube voltage, tube current, kernel--$f_{50}$ pair, slice thickness, pixel size) that enumerates the full $468$-protocol space $\mathcal{A}$($3\times3\times13\times2\times2$). 

During policy training, the per-step reward was the objective $J_p(a_t)$ of the visited protocol, as defined in~\eqref{eq:dose_aware_objective}. For completed patient--protocol combinations, the reward was calculated using the VIT-derived $d'$ value. For missing training combinations, $d'$ was supplied by the random-forest imputation described in Section~\ref{sec:rf}. Thus, a reward was available for every training transition. The discounted return accumulated these rewards over a search budget of $T=8$ steps with a discount factor of $\gamma=0.99$.

A PPO agent \cite{RN29}, implemented with the open-source Stable-Baselines3 library \cite{RN30} (comparable implementations exist in other open-source libraries \cite{RN31,RN32}), learned the policy $\pi_\theta(a\,|\,s)$ and a value baseline, each a multilayer perceptron (a $128$-unit state encoder feeding two $256$-unit hidden layers with ReLU activations). Training used $10^5$ environment steps, learning rate $3\times10^{-4}$, discount factor $\gamma=0.99$, clip range $0.2$, entropy coefficient $0.01$, and an advantage-smoothing coefficient of $0.95$ for generalized advantage estimation (GAE) \cite{gae1506}.

During deployment, for each held-out patient, we roll out the policy 32 times to generate protocol sequences. Because the state transition depends only on the patient embedding, the selected protocol, and the normalized step index, we can generate these rollouts without observing $J_p(a)$ for the held-out patient. We pooled the distinct protocols proposed across the rollouts into a candidate set. The trained random-forest surrogate predicted $d'$ for each candidate protocol, and we calculated the corresponding dose-aware objective using the known protocol-level dose term to rank the candidates.

Once the candidate list is generated, the proposed method can be deployed in both offline and online modes. In the \emph{offline} mode, there is no access to the CT simulations and held-out ground truth values. As such, the top-ranked protocol is used directly. In the \emph{online} mode, we assume we can afford a small CT simulation budget before the diagnostic scan. Thus, up to eight highest-ranked candidates are simulated, and the protocol with the highest \textit{true} objective is used. The two modes are therefore two operating points of one trained policy---trading simulation budget for accuracy---rather than two different methods; we compare them in Section~\ref{sec:results}.

\subsubsection{Baseline Selectors}
\label{sec:baselines}
The exhaustive-search oracle is the upper bound we aim to approximate, not a deployable competitor in practice. To benchmark our PPO agent against a simple one-shot alternative using the same patient representation and surrogate, we compared it with a contextual bandit that ranks the full 468-protocol space using the random-forest surrogate. The bandit treats selection as a one-shot decision: for a given patient, it scores all 468 protocols and commits to the single highest-scoring one with no revisiting. It therefore differs from PPO only in the search strategy. We evaluated both selectors with and without a localizer, with the localizer-blind variants removing the embedding from the input state.

Beyond this learned baseline, we also consider four non-learning-based reference selectors. First, a \emph{random} selection that draws one of the 468 protocols uniformly per patient; because the objective is available across all imaging conditions, its expected recovery was computed in closed form as the mean objective over the 468 protocols divided by the patient-specific oracle, and its empirical distribution is reported. Second, a \emph{cohort-fixed} protocol was selected based on the single protocol that maximized the mean training-cohort objective for every held-out patient (selection on the 51 training patients only), emulating a fixed protocol. Third, a \emph{water-equivalent-diameter (WED) rule} technique was implemented emulating automatic exposure control: WED was computed from each localizer by mapping it to a target tube current \cite{aapm220} via a constant-CNR law $\mathrm{mAs}(\mathrm{WED})=\mathrm{mAs}_{\mathrm{ref}}\,e^{\,k(\mathrm{WED}-\mathrm{WED}_{\mathrm{ref}})}$ (with $k$ fixed to a published value, and in a second variant, calibrated on training patients to remove any systematic dose offset) and assigned to the nearest available dose level in the 468-protocol grid. Because WED selects dose but not reconstruction, it cannot exploit the reconstruction axis of the action space. Therefore, reconstruction was held at the values used in the cohort-fixed protocol. Fourth, a \emph{WED-$k$NN rule} was incorporated to predict the protocol from body size alone: for each held-out patient we computed WED from the localizer, found its $k$ nearest training patients in WED, and applied the protocol that was oracle-optimal for those neighbors (majority vote, with $k$ chosen on the training patients). This isolates body size from the richer localizer content used by the learned selectors.

\subsection{Evaluation and Metrics}

Our primary evaluation metric is objective recovery, i.e.,

\begin{align}
\mathrm{Recovery}_p
=
\frac{J_p(\hat{a}_p)}{J_p(a_p^\star)}.
\label{eq:recovery}
\end{align}

Here $\hat{a}_p$ denotes the protocol generated by the proposed PPO agent (or the baseline selectors), while $a_p^\star$ is the exhaustive-search oracle of Eq.~\ref{eq:oracle}
. Across patients, we aggregated by ratio-of-means---the summed selected objective divided by the summed oracle objective---as the primary metric, because it weights each patient by the magnitude of their oracle objective and prevents near-floor patients, whose small denominators make per-patient ratios volatile, from dominating the average; the unweighted per-patient mean is also reported. We report detectability $d'$ and the selected dose level as secondary metrics.

We repeated all PPO experiments over five random seeds. We report mean recovery across seeds, with seed-level error bars in the corresponding result figures. Patient-level paired tests were used to compare localizer-conditioned PPO against localizer-blind PPO and localizer-conditioned PPO against the localizer-conditioned contextual bandit. Bootstrap confidence intervals across the 12 held-out patients were computed for the main recovery differences.

We report two complementary measures of variability. The seed-level standard deviation (SD) across the five policy seeds captures optimization stability and is shown in the tables and figure error bars. The patient-level 95\% bootstrap confidence interval (CI; 10{,}000 resamples of the 12 held-out patients, averaging the five seeds within each patient before resampling) captures generalization to new patients and is reported in the text and in table and figure captions for the two primary operating points (off-line and on-line).

\section{Results}
\label{sec:results}
\subsection{Exhaustive Search}

\begin{figure}[!t]
\centering
\includegraphics[width=\linewidth]{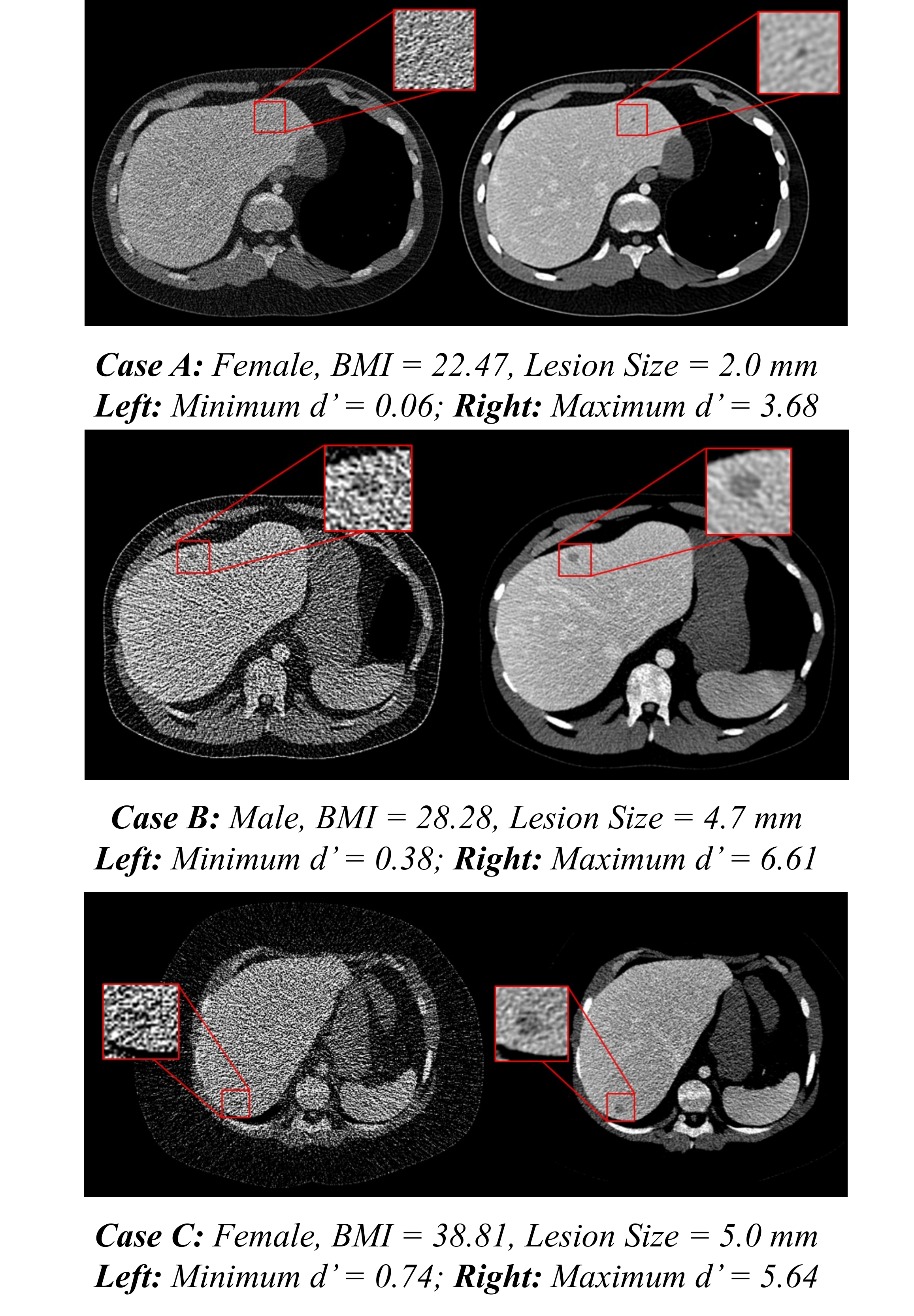}
\caption{Qualitative case studies of protocol impact for three held-out patients spanning body habitus and lesion size. For each patient, the reconstructed axial slice with zoomed lesion inset is shown for the worst protocol in the space (left, minimum $d'$) and the best protocol (right, maximum $d'$). Protocol choice moves the worst-case lesion from nearly invisible (minimum $d'\le0.74$) to clearly detectable (maximum $d'=3.68$--$6.61$), and the detectability-optimal protocol is patient-specific, motivating localizer-conditioned personalized selection.}
\label{fig:casestudy}
\end{figure}

\begin{figure}[!t]
\centering
\includegraphics[width=\columnwidth]{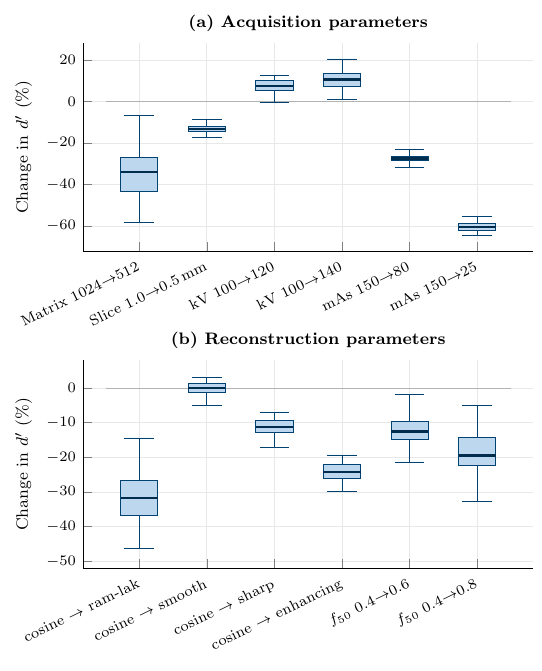}

\caption{Sensitivity of detectability $d'$ to (a) acquisition parameters and (b) reconstruction parameters across the 63-patient cohort. Each box shows the per-patient percent change in $d'$ when one parameter is varied from a common baseline (100~kV, 150~mAs, cosine kernel, $f_{50}=0.4$~mm$^{-1}$, 1.0~mm slices, 1024 matrix); $n=49$--$59$ patients per comparison with both cells simulated. Box: interquartile range across patients; line: median; whiskers: $1.5\times$IQR; outliers omitted.}

\label{fig:landscape}
\end{figure}

\begin{figure}[t]
\centering
\includegraphics[width=\linewidth]{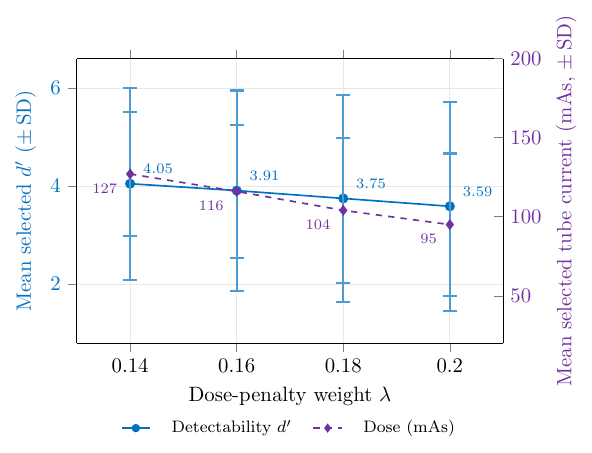}
\caption{Dose--detectability trade-off across the dose-penalty weight $\lambda$, computed from the patient-specific oracle selections of the objective on the 51-patient cohort. Raising $\lambda$ monotonically reduces both the mean selected detectability (left axis, $4.05\rightarrow3.59$) and the mean selected tube current (right axis, $127\rightarrow95$~mAs); error bars: $\pm$1~SD across patients. We set $\lambda=0.18$ as the balanced operating point for the main comparisons.}
\label{fig:lambda_sensitivity}
\end{figure}

Detectability, $d'$, varied widely across the protocol space, confirming that protocol choice impacted lesion visibility. This is illustrated qualitatively in Fig.~\ref{fig:casestudy}: images with the lowest $d'$ values rendered the lesions nearly invisible, whereas the images with the highest $d'$ had improved lesion visibility across patients.

Tube current and reconstruction matrix size caused the largest changes in $d'$: lowering tube current from 150 to 25~mAs reduced $d'$ by $60\pm2\%$ (mean\,$\pm$\,SD across patients), and halving the reconstruction matrix size from 1024 to 512 reduced it by $34\pm14\%$. The reconstruction kernel also reduced $d'$ (e.g., $32\pm8\%$ from the cosine to the ram-lak kernel), whereas slice thickness and $f_{50}$ had smaller effects (Fig.~\ref{fig:landscape}). In contrast, increasing tube voltage from 100 to 120--140~kV increased $d'$ by approximately $10\pm10\%$ and $10\pm13\%$: because the task function fixes lesion contrast, the flux-matched higher-kV spectra reduces image noise without the contrast penalty that affects real lesions. These trends provide an internal physics check on the measurements and motivate patient-specific protocol selection.

The dose-penalty weight, $\lambda$, set the operating point of the objective. Smaller penalties favored image quality and shifted selections toward higher-dose protocols, whereas larger penalties favored dose reduction. We evaluated $\lambda\in\{0.14,0.16,0.18,0.20\}$ using the patient-specific oracle selections of the objective over the 51 training patients. Increasing $\lambda$ monotonically reduced both the mean selected tube current (from $127\pm39$ to $95\pm45$~mAs) and the mean detectability (from $d'=4.05\pm1.96$ to $3.59\pm2.13$; mean\,$\pm$\,SD across patients), tracing a clear dose--detectability trade-off (Fig.~\ref{fig:lambda_sensitivity}). Within the swept range, $\lambda=0.14$ represented an image-quality-priority operating point and $\lambda=0.20$ a dose-priority operating point; we used $\lambda=0.18$ as a balanced operating point.

\subsection{Performance of Random Forest Surrogate}

The random-forest surrogate was evaluated separately from the final oracle-based recovery calculation to ensure that reported performance reflected true VIT-derived objectives. On the $d'$ target, the surrogate achieved Spearman $\rho=0.93$--$0.94$ between predicted and simulated detectability and top-1 protocol-recovery consistency of 0.90--0.94 on held-out patients (the recovery attained by the surrogate's top-ranked protocol relative to the patient-specific oracle). Because selection uses the dose-aware objective $J$ rather than $d'$ alone, we also checked objective-level ranking: the Spearman correlation between predicted and true $J$ over the 468 protocols was 0.90. Top-tail calibration was strong, with Bland--Altman bias of $-0.005$ (95\% limits of agreement $\pm0.10$) and near-zero error within the predicted top decile ($-0.002$). This ranking fidelity supports using the surrogate to fill unsampled training cells and to score candidate protocols at deployment.

\subsection{Performance of the Proposed PPO Agent}

We measured selector performance on held-out patients by calculating objective recovery relative to the patient-specific exhaustive-search oracle. All results in this subsection use the objective at $\lambda=0.18$.

As Table~\ref{tab:patient_level} demonstrates, localizer-conditioned PPO recovered $89.7\%$ (seed SD $0.5$; 95\% CI $80.9$--$95.7$) of the oracle objective by ratio-of-means. To assess how the design of the patient representation contributed to this performance, we conducted three ablation studies. First, compressing the embedding to its leading six principal components (PCA-6) gave a more accurate policy than the full $128$-dimensional embedding ($87.9\pm1.4\%$; PCA-6 ahead by $1.8$~pp, paired $p=0.047$) and reduced seed-to-seed variability. Second, removing the localizer entirely reduced recovery to $79.0\pm0.2\%$---a $10.7$~pp drop (paired 95\% CI $2.9$--$19.5$; exact permutation $p=0.021$). Third, the gain came from how the localizer was encoded: a raw, label-free $768$-dimensional ViT CLS embedding was less accurate than our $d'$-supervised PCA-6 representation, and even matched at six dimensions the label-free embedding trailed by $4.9$~pp---showing that the localizer signal must be distilled by task ($d'$) supervision rather than read from generic features or padded with more dimensions.

A patient-agnostic scorer converges to a single global protocol, whereas the localizer-conditioned scorer preserves per-patient differentiation. Matched patient-representation ablations kept the downstream surrogate and policy architecture fixed while comparing BMI, WED, frozen ViT CLS features, the learned 128-dimensional embedding, and its PCA-reduced variants. The learned PCA-6 representation embedding achieved the highest training-only cross-validated recovery and was therefore fixed for all held-out evaluations.

For reference, a contextual bandit using the same localizer embedding and surrogate reached comparable recovery ($86.4\pm2.1\%$, 95\% CI $78.2$--$91.9$, with the localizer; $82.2\pm3.8\%$ without; five-fold cross-validation). The bandit scores all 468 protocols by surrogate and commits in one shot, whereas PPO commits after scoring only the few protocols its rollouts visit (about 15.3 distinct on average, from 32 rollouts).

\begin{figure}[!t]
\centering
\includegraphics[width=\columnwidth]{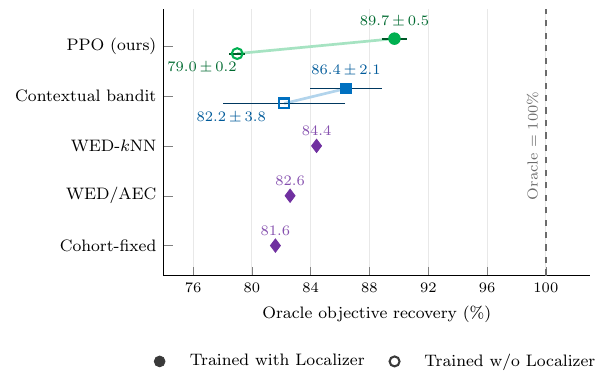}
\caption{Off-line (zero-simulation) oracle objective recovery of the deployable selectors on held-out patients at $\lambda=0.18$; this is the operating point reached with no patient-specific simulation. Filled markers: trained with the localizer; open markers: trained without it; the connector length is the localizer gain ($+10.7$~pp for PPO, $+4.2$~pp for the bandit). Whiskers: $\pm$1~SD (five seeds for PPO, five cross-validation folds for the bandit); the rule-based selectors are deterministic. The patient-level 95\% bootstrap CI for the localizer-conditioned PPO point is 80.9--95.7\%. The exhaustive-search oracle ($100\%$) is shown dashed and the random floor ($31.0\%$) lies off scale.}
\label{fig:recovery}
\end{figure}

Against the non-learned deployable selectors, offline PPO exceeded random selection ($31.0\%$), the cohort-fixed protocol ($81.6\%$), the WED/AEC rule ($82.6\%$), and WED-$k$NN ($84.4\%$) in point estimate, although at $n=12$ the offline margin over the fixed protocol was not statistically significant (paired $p=0.25$). Statistical separation arrived with a small measurement budget: from two simulations onward the policy significantly exceeded the fixed protocol ($p\le0.03$, reaching $p=0.004$ at eight simulations) (Fig.~\ref{fig:recovery}).

\begin{table*}[!t]
\caption{Per-patient results on the 12 held-out test patients ($\lambda=0.18$).}
\label{tab:patient_level}
\centering
\footnotesize
\renewcommand{\arraystretch}{1.15}
\setlength{\tabcolsep}{6pt}
\resizebox{\textwidth}{!}{%
\begin{tabular}{lcc ccc cccc}
\toprule
& \multicolumn{2}{c}{\textbf{Oracle}} & \multicolumn{3}{c}{\textbf{Offline (no simulation)}} & \multicolumn{4}{c}{\textbf{Online (8 simulations)}} \\
\cmidrule(lr){2-3}\cmidrule(lr){4-6}\cmidrule(lr){7-10}
\textbf{XCAT ID} & \textbf{$d'$} & \textbf{mAs} & \textbf{$d'$} & \textbf{mAs} & \textbf{Rec.\ (\%)} & \textbf{$d'$} & \textbf{mAs} & \textbf{Rec.\ (\%)} & \textbf{VIT budgets} \\
\midrule
pt152 & 7.93 & 150 & 7.92\,$\pm$\,0.01 & 150            & 99.9\,$\pm$\,0.2 & 7.92\,$\pm$\,0.01 & 150            & 99.9\,$\pm$\,0.2  & 5.6 \\
pt300 & 6.62 & 150 & 4.95\,$\pm$\,0.03 & 80             & 87.5\,$\pm$\,0.6 & 6.62\,$\pm$\,0.00 & 150            & 100.0\,$\pm$\,0.1 & 8.0 \\
pt269 & 5.13 & 150 & 4.86\,$\pm$\,0.01 & 150            & 96.6\,$\pm$\,4.1 & 5.03\,$\pm$\,0.13 & 150            & 96.7\,$\pm$\,4.0  & 6.8 \\
pt157 & 4.58 & 150 & 2.70\,$\pm$\,0.01 & 80             & 70.2\,$\pm$\,0.2 & 4.34\,$\pm$\,0.48 & 136\,$\pm$\,28 & 99.1\,$\pm$\,1.8  & 7.2 \\
pt141 & 4.46 & 150 & 3.17              & 80             & 92.0            & 3.68\,$\pm$\,0.61 & 108\,$\pm$\,34 & 95.1\,$\pm$\,3.1  & 8.0 \\
pt279 & 4.34 & 150 & 4.34\, & 150 & 100\ & 4.34\,$\pm$\,0.00 & 150            & 100.0\,$\pm$\,0.0 & 7.8 \\
pt200 & 3.73 & 150 & 1.85              & 80             & 55.8\,$\pm$\,0.1 & 3.39\,$\pm$\,0.59 & 136\,$\pm$\,28 & 93.2\,$\pm$\,9.3  & 7.8 \\
pt252 & 3.34 & 80  & 3.29\,$\pm$\,0.08 & 80             & 99.0 & 3.34\,$\pm$\,0.00 & 80             & 100.0\,$\pm$\,0.0 & 8.0 \\
pt143 & 2.87 & 80  & 2.86 & 80 & 98.8 & 2.86\,$\pm$\,0.03 & 80             & 99.1\,$\pm$\,1.6  & 7.4 \\
pt154 & 1.52 & 25  & 2.07              & 80             & 80.1            & 1.63\,$\pm$\,0.22 & 36\,$\pm$\,22  & 96.0\,$\pm$\,7.9  & 8.0 \\
pt167 & 1.40 & 25  & 2.18\,$\pm$\,0.02 & 80             & 93.7 & 2.20\,$\pm$\,0.00 & 80             & 96.3\,$\pm$\,0.0  & 7.8 \\
pt163 & 0.88 & 25  & 1.37\,$\pm$\,0.01 & 80             & 58.4\,$\pm$\,1.3 & 0.82\,$\pm$\,0.07 & 25             & 92.9\,$\pm$\,8.5  & 8.0 \\
\midrule
Mean  & 3.90 & 107 & 3.47\,$\pm$\,0.04 & 99\,$\pm$\,2   & 89.7\,$\pm$\,0.5 & 3.85\,$\pm$\,0.11 & 106.8\,$\pm$\,4.8 & 98.2\,$\pm$\,1.5 & 7.5 \\
\bottomrule
\end{tabular}}

\vspace{2pt}
{\footnotesize\begin{minipage}{\textwidth}\raggedright
\textit{Notes.} ``Oracle'' is the exhaustive-search optimum. ``Off-line'' commits from the surrogate with no patient-specific simulation; ``On-line'' simulates up to eight surrogate-ranked candidates and commits the best by true objective. ``VIT budgets'' is the mean number of distinct protocols actually simulated over five seeds; it falls below eight when the policy proposes fewer than eight distinct candidates. Values are mean\,$\pm$\,SD over five seeds (entries without a spread were identical across all five). For the near-floor lesions (pt154, pt167, pt163) the oracle uses the minimum tube current, so PPO's raw $d'$ can exceed the oracle's while recovery stays below 100\%. The ``Mean'' recovery is the ratio-of-means over the 12 patients (off-line $89.7\pm0.5\%$, patient-level 95\% CI 80.9--95.7; on-line $98.2\pm1.5\%$, 95\% CI 96.6--99.2); the unweighted per-patient means are $86.1\pm0.5\%$ and $97.4\pm1.8\%$. Per-patient spreads are seed-level SD.
\end{minipage}}
\end{table*}

When a small simulation budget is available before the diagnostic scan, the policy's candidates can be verified by simulation (Table~\ref{tab:efficiency}): simulating up to the eight highest surrogate-ranked candidates and committing the best by measured objective recovered $98.2\%$ (seed SD $1.5$; 95\% CI $96.6$--$99.2$) of the oracle, while reducing the number of required simulations by 98.3\%. 

Because both modes draw from the same candidate list, the offline point is the zero-simulation operating point of this curve. Both modes share one candidate list: the policy is rolled out 32 times, and the distinct protocols it proposes (about $15.3$ on average, $3.3\%$ of the grid) are ranked by surrogate score. The offline mode commits the top-ranked candidate ($89.7\%$). 

For some patients, repeated PPO rollouts proposed overlapping protocols, yielding fewer than eight distinct candidates. Consequently, fewer than eight protocols were simulated for those patients, and the cohort mean was 7.5 simulations per patient. Random search (best-of-$k$) over the same budget reached only $64.8\pm5.0\%$ at eight evaluations, and the contextual bandit, which scores all 468 protocols with the surrogate, tracked below PPO at every budget. At the eight-simulation budget, the committed protocols averaged $106.8$~mAs, within $0.3$~mAs of the oracle's mean tube current ($107.1$~mAs), showing that the policy matched the oracle's dose level rather than defaulting to maximum tube current.

\subsection{Patient-Level Results and Selected-Protocol Trade-offs}

\begin{table}[!t]
\caption{Oracle recovery versus on-line VIT budget ($\lambda=0.18$).}
\label{tab:efficiency}
\centering
\footnotesize
\renewcommand{\arraystretch}{1.2}
\setlength{\tabcolsep}{6pt}
\begin{tabular}{cccc}
\toprule
\textbf{Simulations} & \textbf{PPO (ours)} & \textbf{Bandit} & \textbf{Random} \\
\textbf{per patient} & (\%) & (\%) & (\%) \\
\midrule
0 (off-line) & \textbf{89.7\,$\pm$\,0.5} & 86.4\,$\pm$\,2.1 & 31.0 \\
2            & 90.5\,$\pm$\,0.7 & 87.5\,$\pm$\,1.5 & 43.9  \\
4            & 94.0\,$\pm$\,0.4 & 92.0\,$\pm$\,1.5 & 55.1  \\
6            & 97.0\,$\pm$\,0.8 & 94.9\,$\pm$\,1.3 & 60.9  \\
8            & \textbf{98.2\,$\pm$\,1.5} & 95.8\,$\pm$\,1.3 & 64.8\\
\midrule
\multicolumn{4}{@{}l}{\textit{Single-protocol baselines} (no simulation)}\\
WED-$k$NN        & \multicolumn{3}{c}{84.4} \\
WED\,/\,AEC      & \multicolumn{3}{c}{82.6} \\
Cohort-fixed     & \multicolumn{3}{c}{81.6} \\
Random           & \multicolumn{3}{c}{31.0} \\
\midrule
Exhaustive (468) & \multicolumn{3}{c}{100.0} \\
\bottomrule
\end{tabular}

\vspace{2pt}
{\footnotesize\begin{minipage}{\columnwidth}\raggedright
\textit{Notes.} Recovery is ratio-of-means over the 12 held-out patients. Tabulated values are the seed-level mean\,$\pm$\,SD (PPO over five seeds, bandit over five folds); for localizer-conditioned PPO the patient-level 95\% bootstrap CI is 80.9--95.7 at the off-line point ($k{=}0$) and 96.6--99.2 at $k{=}8$. At $k$ simulations, each learned selector simulates its $k$ top surrogate-ranked candidates and commits the best by true objective; row ``0'' is the off-line mode. Random simulates $k$ uniform draws. Single-protocol baselines and the exhaustive reference do not need candidate ranking.
\end{minipage}}
\end{table}

Given the pronounced inter-patient variability within the held-out cohort, we report protocol selections and recovery on a per-patient basis (Table~\ref{tab:patient_level}). The table reports absolute detectability and tube current at each selected protocol. Across the 12 patients, localizer-conditioned PPO exceeded the localizer-blind PPO policy by 10.7~pp of oracle recovery (95\% bootstrap CI $[2.9,\,19.5]$; exact  permutation $p=0.021$), and the PCA-6 embedding exceeded the full 128-dimensional embedding by 1.8~pp ($[0.4,\,3.1]$; $p=0.047$); both benefits were statistically significant. PPO's point estimate also exceeded the cross-validated contextual bandit by 3.3~pp and the strongest size-based rule, WED-$k$NN, by 5.3~pp. The localizer benefit was concentrated in lower-oracle or near-floor patients, where the localizer-blind policy was more likely to choose low-detectability protocols.

\begin{table*}[!t]
\caption{Per-patient selected protocols (tube voltage / tube current / kernel) and recovery, offline versus online, at $\lambda=0.18$.}
\label{tab:patient_protocol}
\centering
\footnotesize
\renewcommand{\arraystretch}{1.2}
\setlength{\tabcolsep}{7pt}
\begin{tabular}{l ccc ccc ccc c}
\toprule
& \multicolumn{3}{c}{\textbf{Oracle}} & \multicolumn{3}{c}{\textbf{Offline (no simulation)}} & \multicolumn{3}{c}{\textbf{Online (8 simulations)}} & \\
\cmidrule(lr){2-4}\cmidrule(lr){5-7}\cmidrule(lr){8-10}
\textbf{XCAT ID} & kV & mAs & kernel & kV & mAs & kernel & kV & mAs & kernel & \textbf{Rec. off$\rightarrow$on (\%)} \\
\midrule
pt152 & 120 & 150 & cos & 120 & 150 & cos & 120 & 150 & cos & 99.9\,$\rightarrow$\,99.9 \\
pt300 & 120 & 150 & cos & 140 & 80  & smo & \textbf{120} & \textbf{150} & \textbf{cos} & 87.5\,$\rightarrow$\,100.0 \\
pt269 & 140 & 150 & smo & 120 & 150 & smo & \textbf{140} & 150 & smo & 96.6\,$\rightarrow$\,96.7 \\
pt157 & 120 & 150 & smo & 140 & 80  & smo & \textbf{120} & \textbf{150} & smo & 70.2\,$\rightarrow$\,99.1 \\
pt141 & 140 & 150 & cos & 140 & 80  & smo & 140 & 80  & \textbf{cos} & 92.0\,$\rightarrow$\,95.1 \\
pt279 & 140 & 150 & smo & 140 & 150 & smo & 140 & 150 & smo & 100.0\,$\rightarrow$\,100.0 \\
pt200 & 140 & 150 & cos & 140 & 80  & smo & 140 & \textbf{150} & \textbf{cos} & 55.8\,$\rightarrow$\,93.2 \\
pt252 & 140 & 80  & smo & 140 & 80  & smo & 140 & 80  & smo & 99.0\,$\rightarrow$\,100.0 \\
pt143 & 140 & 80  & cos & 120 & 80  & cos & \textbf{140} & 80  & cos & 98.8\,$\rightarrow$\,99.1 \\
pt154 & 140 & 25  & cos & 140 & 80  & smo & 140 & \textbf{25}  & \textbf{cos} & 80.1\,$\rightarrow$\,96.0 \\
pt167 & 140 & 25  & cos & 140 & 80  & cos & 140 & 80  & \textbf{smo} & 93.7\,$\rightarrow$\,96.3 \\
pt163 & 120 & 25  & smo & 140 & 80  & cos & \textbf{120} & \textbf{25}  & \textbf{smo} & 58.4\,$\rightarrow$\,92.9 \\
\midrule
\end{tabular}

\vspace{2pt}
{\footnotesize\begin{minipage}{\textwidth}\raggedright
\textit{Notes.} Each entry is the majority (modal) protocol across the five seeds. \textbf{Bold} online values are those for which the eight-simulation budget changed relative to the offline choice; comparing the online block with the oracle shows whether the change reached the optimum. Kernel: cos = cosine, smo = smooth. Recovery is ratio-of-means over the five seeds. Slice thickness ($1.0$~mm), pixel size ($1024$ matrix), and $f_{50}$ ($0.4$~mm$^{-1}$) were identical for every selection and every oracle and are omitted.
\end{minipage}}
\end{table*}

Removing the localizer reduced the performance because in that scenario surrogate depended only on the protocol and ranked protocols the same way for every patient; its argmax therefore converged to a single globally highest-scoring protocol and the selector degenerated to a fixed protocol---its selected tube current was fixed at 150~mAs for every patient (a $+88\%$ dose increase) with near-zero seed variability (SD $0.22$). The localizer-conditioned surrogate instead kept a per-patient ranking, making the selection patient-specific rather than one-size-fits-all.

With a budget of eight simulations per patient, the same policy reached near-oracle recovery for every patient (Table~\ref{tab:patient_level}, online columns). The three patients that had the weakest performance in the offline mode had improved the most---pt157 ($70.2\rightarrow99.1\%$), pt200 ($55.8\rightarrow93.2\%$), and pt163 ($58.4\rightarrow92.9\%$)---and three patients (pt300, pt279, pt252) reached the exact oracle protocol. Notably, these weakest off-line cases were the largest patients: off-line recovery correlated strongly and negatively with the water-equivalent diameter (WED) ($\rho=-0.83$, $p=0.002$, $n=12$), and the three lowest off-line recoveries belonged to the three largest patients by WED. The eight-simulation budget concentrated its benefit on exactly these patients (WED versus
per-patient recovery gain, $\rho=+0.84$, $p=0.001$), narrowing the gap between the six largest- and six smallest-WED patients from $21.7$ percentage points off-line ($75.3\%$ vs.\ $97.0\%$) to $3.8$ points on-line. By contrast, neither the deviation of a patient's oracle protocol from the cohort-modal protocol (Hamming distance, $\rho=-0.40$, $p=0.20$) nor the sharpness of the objective landscape ($\rho=+0.22$, $p=0.50$) was associated with off-line difficulty, indicating that patient size---rather than an idiosyncratic optimum---was the dominant driver. The committed tube current averaged $106.8$~mAs, close to the oracle's $107.1$~mAs, so the policy matched the oracle's dose rather than defaulting to the maximum.

Compared to the offline mode, the online mode with 8 simulations narrowed the patient-level 95\% confidence interval from 14.9 to 2.6 percentage points: offline recovery ranged from 55.8\% to 100.0\% across patients, whereas after eight simulations every patient fell between 92.9\% and 100\%. The budget therefore reduced not only the mean shortfall but also the patient-to-patient variability of the committed protocols. Table~\ref{tab:patient_protocol} shows, per patient, the protocol the policy committed offline and online, next to the oracle. Across the eight simulations, the tube current or kernel changed for most patients, moving each dose toward the oracle in the right direction.

\section{Discussion}

This study proposed a framework that combines virtual imaging trials with reinforcement learning, enabling efficient and accurate CT protocol optimization. The virtual imaging framework served as a forward model to create a patient-specific protocol landscape and exhaustive-search oracle over a 468-protocol acquisition--reconstruction space. This oracle was approximated by training a PPO policy using only CT localizer image and a small budget of candidate protocols.

The measured protocol landscape followed known CT physics. Higher tube current improved detectability ($d'$) by suppressing image noise. Because the task function had a fixed lesion contrast, higher tube voltage modestly improved $d'$ through noise reduction caused by higher flux at the detector due to a more penetrating beam at higher tube potentials; for actual lesions, lower tube potential may still increase lesion-to-parenchyma contrast, making the tube voltage trend task-dependent. Softer kernels and lower filter cutoffs improved low-contrast detectability by reducing noise texture, despite the associated loss in spatial resolution. These trends are consistent with established CT image-quality behavior and prior studies showing that dose, tube potential, and reconstruction kernel can strongly affect contrast-to-noise ratio and task-based detectability in low-contrast liver lesion imaging \cite{RN25,RN17}. 

The dose-aware objective changed the problem from pure image-quality maximization to a detectability-dose trade-off. Pure \(d'\)-maximization tend to favor high-dose settings, especially for subtle low-contrast lesions. The dose penalty rewards higher detectability only when the gain justifies the added tube current. The parameter \(\lambda\) therefore defines an operating point on a dose--detectability frontier. In this study, \(\lambda=0.18\) was used as a balanced operating point rather than a universal clinical optimum; in practice, different \(\lambda\) values may be selected according to dose constraints and imaging task priorities.

We introduced a single policy that operates in two inexpensive modes. Off-line, with no patient-specific simulation, it recovered $89.7\%$ (seed SD $0.5$; 95\% CI $80.9$--$95.7$) of the exhaustive-search oracle from surrogate scoring alone---above every deployable rule we tested ($81.6$--$84.4\%$). On-line, simulating up to eight candidate protocols raised recovery to $98.2\%$ (seed SD $1.5$; 95\% CI $96.6$--$99.2$), matching the 468-protocol oracle at under $2\%$ of its evaluation cost. This target is not easy to reach by chance: only $0.5\%$ of protocols fall within $5\%$ of the oracle, and random search (best-of-$k$) reached just $64.8\pm5.0\%$ at the same eight-simulation budget.

At this action-space size, a contextual bandit that scores all 468 protocols with the same surrogate performed comparably. In comparison, PPO had two advantages. First, its candidate set is bounded: even its worst-ranked candidate recovered $44.5\%$ of the oracle, whereas the full grid contains protocols recovering as low as $-23.9\%$(recovery is negative when a protocol's dose penalty exceeds its normalized detectability, giving $J<0$); a surrogate error can therefore lead a full-grid selector to an inferior protocol but cannot do so for the policy. Second, enumerating and ranking the full grid is feasible for 468 protocols but not for the near-continuous protocol spaces of modern scanners, where a policy that proposes a small candidate set scales with the search budget rather than with the size of the protocol space. 

This study had several limitations. First, we evaluated the proposed framework using a limited number of human models with liver lesions. Expanding the patient cohort may better characterize inter-patient variability and improve generalizability. Second, this study evaluated the proposed CT optimization framework in the context of liver lesion detection. Given that our proposed approach can be readily extended to other lesion types and diagnostic objectives, future studies may investigate its performance and utility across other imaging applications. Third, we represented the dose term with a protocol-level tube-current surrogate rather than size-specific dose estimates or organ dosimetry. These metrics may be more appropriate for certain imaging tasks and may be used in future studies. Fourth, the action space was finite and did not include some other acquisition settings including automatic tube-current modulation, pitch, iterative reconstruction, or deep-learning reconstruction. Fifth, within the action space we did explore, the selected and oracle protocols never varied three reconstruction parameters—slice thickness, pixel size, and filter cutoff ($f_{50}$), which remained fixed across every patient and every oracle. Therefore, the effective decision space was lower-dimensional than the nominal 468 protocols. This indicates that our search space was not the most efficient choice and could be adaptively pruned in future work; at the same time, the optimizer's consistent avoidance of the alternative settings is consistent with established CT physics (e.g., low-contrast detection favoring lower image noise and a lower filter cutoff) and provides a data-driven basis for compacting the space. Finally, the localizer images, scanner model, and reconstruction pipeline were generated within a virtual imaging setting; prospective evaluation with clinical localizers, vendor-specific acquisition models, and reader-calibrated task thresholds \cite{RN18} will be required before clinical translation.

\section{Conclusion}

This study established a framework that integrates virtual imaging trials with reinforcement learning to optimize CT protocols without a need for exhaustive search. Under the studied dose-aware objective, a single localizer-conditioned policy recovered $89.7\%$ (seed SD $0.5$; 95\% CI $80.9$--$95.7$) of the exhaustive-search oracle with no patient-specific simulation and $98.2\%$ (seed SD $1.5$; 95\% CI $96.6$--$99.2$) when up to eight protocols per patient could be simulated on-line. These results show that a protocol-selection policy can be trained on virtual imaging data and applied prospectively to a new patient from their localizer image alone efficiently.

\bibliographystyle{ieeetr}
\bibliography{RL_VIT_2}

\begin{thebibliography}{10}

\bibitem{RN1}
D.~D. Cody, T.~S. Fisher, D.~A. Gress, R.~R. Layman, M.~F. McNitt-Gray, R.~J.
  Pizzutiello, and L.~A. Fairobent, ``{AAPM} medical physics practice guideline
  1.a: {CT} protocol management and review practice guideline,'' {\em Journal
  of Applied Clinical Medical Physics}, vol.~14, no.~5, pp.~3--12, 2013.

\bibitem{RN2}
K.~Martini, J.~W. Moon, M.~P. Revel, S.~Dangeard, C.~Ruan, and G.~Chassagnon,
  ``Optimization of acquisition parameters for reduced-dose thoracic {CT}: A
  phantom study,'' {\em Diagn Interv Imaging}, vol.~101, no.~5, pp.~269--279,
  2020.

\bibitem{RN3}
F.~Zarb, L.~Rainford, and M.~F. McEntee, ``Developing optimized {CT} scan
  protocols: phantom measurements of image quality,'' {\em Radiography},
  vol.~17, no.~2, pp.~109--114, 2011.

\bibitem{RN14}
E.~Abadi {\em et~al.}, ``Virtual clinical trials in medical imaging: a
  review,'' {\em J Med Imaging (Bellingham)}, vol.~7, no.~4, p.~042805, 2020.

\bibitem{RN11}
B.~Barufaldi {\em et~al.}, ``Virtual clinical trials in medical imaging system
  evaluation and optimisation,'' {\em Radiat Prot Dosimetry}, vol.~195,
  no.~3-4, pp.~363--371, 2021.

\bibitem{RN12}
L.~Vancoillie {\em et~al.}, ``Optimized signal of calcifications in wide-angle
  digital breast tomosynthesis: a virtual imaging trial,'' {\em Eur Radiol},
  2024.

\bibitem{RN13}
E.~Abadi, W.~P. Segars, B.~Harrawood, S.~Sharma, A.~Kapadia, and E.~Samei,
  ``Virtual clinical trial for quantifying the effects of beam collimation and
  pitch on image quality in computed tomography,'' {\em J Med Imaging
  (Bellingham)}, vol.~7, no.~4, p.~042806, 2020.

\bibitem{RN15}
E.~Abadi, W.~P. Segars, N.~Felice, S.~Sotoudeh-Paima, E.~A. Hoffman, X.~Wang,
  W.~Wang, D.~Clark, S.~Ye, G.~Jadick, {\em et~al.}, ``{AAPM} truth-based {CT}
  (truect) reconstruction grand challenge,'' {\em Med Phys}, vol.~52, no.~4,
  pp.~1978--1990, 2025.

\bibitem{RN16}
T.~J. Sauer, A.~Bejan, P.~Segars, and E.~Samei, ``Development and {CT}
  image-domain validation of a computational lung lesion model for use in
  virtual imaging trials,'' {\em Med Phys}, vol.~50, no.~7, pp.~4366--4378,
  2023.

\bibitem{RN4}
E.~Abadi, B.~Harrawood, S.~Sharma, A.~Kapadia, W.~P. Segars, and E.~Samei,
  ``Dukesim: A realistic, rapid, and scanner-specific simulation framework in
  computed tomography,'' {\em IEEE Trans Med Imaging}, vol.~38, no.~6,
  pp.~1457--1465, 2019.

\bibitem{RN19}
C.~McCabe, B.~Harrawood, E.~Samei, and E.~Abadi, ``In silico modeling of a
  clinical photon-counting {CT} system: verification and validation,'' {\em
  Medical Physics}, vol.~52, no.~6, pp.~3840--3853, 2025.

\bibitem{RN20}
E.~Abadi {\em et~al.}, ``Development of a scanner-specific simulation framework
  for photon-counting computed tomography,'' {\em Biomed Phys Eng Express},
  vol.~5, no.~5, 2019.

\bibitem{RN21}
S.~S. Shankar {\em et~al.}, ``Task-based validation and application of a
  scanner-specific {CT} simulator using an anthropomorphic phantom,'' {\em
  Medical Physics}, vol.~49, no.~12, pp.~7447--7457, 2022.

\bibitem{RN27}
J.~Wells {\em et~al.}, ``Announcing supercomputer summit,'' 2016.

\bibitem{RN5}
D.~P. Clark and C.~T. Badea, ``{MCR} toolkit: a {GPU}-based toolkit for
  multi-channel reconstruction of preclinical and clinical x-ray {CT} data,''
  {\em Med Phys}, vol.~50, no.~8, pp.~4775--4796, 2023.

\bibitem{RN26}
T.~B. Smith, J.~Solomon, and E.~Samei, ``Estimating detectability index in
  vivo: development and validation of an automated methodology,'' {\em Journal
  of Medical Imaging}, vol.~5, no.~3, p.~031403, 2017.

\bibitem{RN22}
J.~Sanders, L.~Hurwitz, and E.~Samei, ``Patient-specific quantification of
  image quality: an automated method for measuring spatial resolution in
  clinical {CT} images,'' {\em Medical Physics}, vol.~43, no.~10,
  pp.~5330--5338, 2016.

\bibitem{RN23}
J.~G. Ott, F.~Becce, P.~Monnin, S.~Schmidt, F.~O. Bochud, and F.~R. Verdun,
  ``Update on the non-prewhitening model observer in computed tomography for
  the assessment of the adaptive statistical and model-based iterative
  reconstruction algorithms,'' {\em Physics in Medicine \& Biology}, vol.~59,
  no.~15, p.~4047, 2014.

\bibitem{RN24}
O.~Christianson, J.~Winslow, D.~P. Frush, and E.~Samei, ``Automated technique
  to measure noise in clinical {CT} examinations,'' {\em American Journal of
  Roentgenology}, vol.~205, no.~1, pp.~W93--W99, 2015.

\bibitem{dosovitskiy2021vit}
A.~Dosovitskiy, L.~Beyer, A.~Kolesnikov, D.~Weissenborn, X.~Zhai,
  T.~Unterthiner, M.~Dehghani, M.~Minderer, G.~Heigold, S.~Gelly, J.~Uszkoreit,
  and N.~Houlsby, ``An image is worth 16x16 words: Transformers for image
  recognition at scale,'' {\em arXiv preprint arXiv:2010.11929}, 2020.

\bibitem{RN29}
J.~Schulman, F.~Wolski, P.~Dhariwal, A.~Radford, and O.~Klimov, ``Proximal
  policy optimization algorithms,'' {\em arXiv preprint arXiv:1707.06347},
  2017.

\bibitem{RN30}
A.~Raffin, A.~Hill, A.~Gleave, A.~Kanervisto, M.~Ernestus, and N.~Dormann,
  ``Stable-baselines3: reliable reinforcement learning implementations,'' {\em
  Journal of machine learning research}, vol.~22, no.~268, pp.~1--8, 2021.

\bibitem{RN31}
A.~Bou {\em et~al.}, ``Torchrl: a data-driven decision-making library for
  pytorch,'' {\em arXiv preprint arXiv:2306.00577}, 2023.

\bibitem{RN32}
A.~Serrano-Munoz, D.~Chrysostomou, S.~B\o{}gh, and N.~Arana-Arexolaleiba,
  ``skrl: modular and flexible library for reinforcement learning,'' {\em
  Journal of Machine Learning Research}, vol.~24, no.~254, pp.~1--9, 2023.

\bibitem{gae1506}
J.~Schulman, P.~Moritz, S.~Levine, M.~Jordan, and P.~Abbeel, ``High-dimensional
  continuous control using generalized advantage estimation,'' in {\em Proc.
  Int. Conf. Learn. Representations (ICLR)}, 2016.

\bibitem{aapm220}
{American Association of Physicists in Medicine}, ``Use of water equivalent
  diameter for calculating patient size and size-specific dose estimates
  ({SSDE}) in {CT},'' Tech. Rep. AAPM Report No. 220, AAPM Task Group 220,
  2014.

\bibitem{RN25}
N.~Felice {\em et~al.}, ``Photon-counting computed tomography versus
  energy-integrating computed tomography for detection of small liver lesions:
  comparison using a virtual framework imaging,'' {\em Journal of Medical
  Imaging}, vol.~11, no.~5, p.~053502, 2024.

\bibitem{RN17}
S.~T. Schindera {\em et~al.}, ``Hypervascular liver tumors: low tube voltage,
  high tube current multi--detector row {CT} for enhanced detection---phantom
  study,'' {\em Radiology}, vol.~246, no.~1, pp.~125--132, 2008.

\bibitem{RN18}
J.~Solomon and E.~Samei, ``Correlation between human detection accuracy and
  observer model-based image quality metrics in computed tomography,'' {\em
  Journal of Medical Imaging}, vol.~3, no.~3, p.~035506, 2016.

\end{thebibliography}

\end{document}